\documentclass{article}
\usepackage{jfrExamplee}
\usepackage{apalike}
\usepackage{setspace}
\usepackage{booktabs} 
\usepackage{scalefnt} 
\usepackage{multirow} 
\usepackage{graphicx} 
\usepackage{hyperref} 
\usepackage[skip=0pt,singlelinecheck=off]{subcaption}

\providecommand{\keywords}[1]{\textbf{\textit{Keywords: }} #1}
\newcommand*{\doi}[1]{\href{http://doi.org/\detokenize{#1}}{doi:\detokenize{#1}}}

\title{Two-Step Meta-Learning for Time-Series Forecasting Ensemble~\thanks{Version~\today \hfill Accepted to \href{https://ieeexplore.ieee.org/document/9410467}{IEEE Access} journal in April 22, 2021}}

\author{
Evaldas Vaiciukynas \\
Department of Information Systems \\
Faculty of Informatics \\
Kaunas University of Technology \\
Kaunas, Lithuania \\
\texttt{evaldas.vaiciukynas@ktu.lt} \\
\And
Paulius Danenas \\
Centre of Information Systems Design Technologies \\
Faculty of Informatics \\
Kaunas University of Technology \\
Kaunas, Lithuania \\
\texttt{paulius.danenas@ktu.lt} \\
\AND
Vilius Kontrimas \\
Private limited liability company "Rivile" \\
Vilnius, Lithuania \\
\texttt{vilius.kontrimas@advantes.tech} \\
\And
Rimantas Butleris \\
Centre of Information Systems Design Technologies \\
Faculty of Informatics \\
Kaunas University of Technology \\
Kaunas, Lithuania \\
\texttt{rimantas.butleris@ktu.lt} \\
}

\begin{document}

\maketitle

\begin{abstract}
Amounts of historical data collected increase and business intelligence applicability with automatic forecasting of time series are in high demand. While no single time series modeling method is universal to all types of dynamics, forecasting using an ensemble of several methods is often seen as a compromise. Instead of fixing ensemble diversity and size, we propose to predict these aspects adaptively using meta-learning. Meta-learning here considers two separate random forest regression models, built on 390 time-series features, to rank 22 univariate forecasting methods and recommend ensemble size. The forecasting ensemble is consequently formed from methods ranked as the best, and forecasts are pooled using either simple or weighted average (with a weight corresponding to reciprocal rank). The proposed approach was tested on 12561 micro-economic time-series (expanded to 38633 for various forecasting horizons) of M4 competition where meta-learning outperformed Theta and Comb benchmarks by relative forecasting errors for all data types and horizons. Best overall results were achieved by weighted pooling with a symmetric mean absolute percentage error of 9.21\% versus 11.05\% obtained using the Theta method.
\end{abstract}


\section{Introduction}
\label{sec:introduction}
Forecasting key performance indicators and any essential dynamics for an organization should be a high-priority business intelligence task. The aim is to envisage indicator values into the future, based on historical observations. An accurate forecast mitigates uncertainties about the future outlook and can reduce errors in decisions and planning, directly influencing the achievability of goals and contributing to risk management. Forecasting should be an integral part of the decision-making activities in management~\cite{Hyndman2010} since the strategic success of the organization depends upon the practical relation between accuracy of forecast and flexibility of resource allocation plan~\cite{Wacker2002}. It is expected that increasing amounts of historical data records, which constitute valuable resources for the forecasting task, will facilitate accurate forecasting and boost these forecasts' importance. Although \cite{Petropoulos2014}, while researching main determinants of forecasting accuracy, found that increasing time-series length has a small positive effect on forecasting accuracy, which contradicts insights from the machine learning and deep learning fields.

Almost 40 years ago worldwide, Makridakis forecasting competitions had started (organized in 1982, 1993, 2000, 2018, and 2020) with a goal to benchmark progress in forecasting techniques and derive scientific insights in time-series forecasting. During these events, teams of participants compete to obtain forecasts for ever-increasing amounts of time-series from diverse fields. Results are summarized into recommendations on the usefulness of various time-series models or their ensembles. In the recent M4 competition~\cite{Makridakis2018} the leading forecasting techniques (12 from 17 most accurate ones) featured model ensembles that pool forecasts of several, mainly statistical, models. The best solution was submitted by Uber Technologies, where the hybrid technique combined a statistical forecasting model with neural network architecture. The next most successful submission~\cite{Manso2020} featured an ensemble of statistical models where weights were thoroughly tuned, and the machine learning model learned these weight recommendations for later prediction. Insights after an older M3 competition~\cite{Makridakis2000} were that forecasts from univariate time-series models almost always (except for annual data) are more accurate than forecasts from multivariate time-series models with external variables (i.e., macroeconomic indicators). Comparison between univariate approaches revealed that more complex models do not guarantee higher accuracy.

Proceeding from results of Makridakis competitions~\cite{Makridakis2018,Makridakis2000} and numerous academic researches~\cite{Clemen1989,Hansen1990,Hendry2004,Timmermann2006,Kolassa2011} it can be concluded that an ensemble of univariate time-series models often outperforms the best member of the ensemble with respect to forecasting accuracy. The success of forecasting ensemble lies in the diversity of its members~\cite{Oliveira2015}, which contributes to robustness against concept drift~\cite{Zang2014} and enhances algorithmic stability~\cite{Zou2004}. Besides ensemble diversity, the individual accuracy of its members is also of utmost importance~\cite{Lemke2010}. A simple arithmetic average with all members weighted equally often outperforms more complex strategies for combining forecasts from several models. Advanced strategies seek to find optimal weights, for example, based on the model's Akaike information criterion~\cite{Kolassa2011} or model's in-sample forecasting errors~\cite{Smith2009} on the last dynamics of time-series in question. In practice, to avoid corrupting the final forecast by a single inaccurate model, variants of robust average or simply median are recommended in forecast pooling~\cite{Hendry2004}. Choice of weights for ensemble members often relies upon in-sample forecasting errors. However, when the approximate ranking of the model pool is available instead of exact errors, the weight could be derived from the model's reciprocal rank~\cite{Aiolfi2006}.

There exists no forecasting method that performs best on all types of time-series~\cite{Bauer2020}. However, efforts to create more universal approaches seek to adjust ensemble size and choose potential members or weights adaptively based on dynamics we try to extrapolate into the future. The argument that the relative accuracy of forecasting methods depends upon the properties of the time-series and information on dynamics at hand can be exploited to choose a suitable model is an old one~\cite{Reid1972}.

Our research explores an adaptive construction of a forecasting ensemble consisting of various statistical and a few machine learning methods with a meta-learning approach. Meta-learners here seek to rank a pool of methods and recommend ensemble size based on historical time-series data characteristics. Recommendations of introduced forecasting assistants are based on training regression meta-models through forecasting experiments on a diverse set of real-world examples - micro-economic time-series from M4 competition. Experiments compare introduced forecasting ensemble based on recommendations from assistants with the best benchmark methods from M4 competition - Theta and Comb, which were outperformed only by 17 out of 49 submissions in M4 competition~\cite{Makridakis2018,Makridakis2020}.

\section{Related work}
\label{sec:related}

The usefulness of statistics summarizing the data available in a time-series in predicting the relative accuracy of different forecasting methods was explored in~\cite{Maede2000} where they created regression models to predict the expected error of the forecasting method for the time-series at hand. More similar research to our idea of forecasting assistants, after early expert system with rules derived by human analysts in~\cite{Collopy1992}, are forecasting techniques based on meta-learning~\cite{Lemke2010,Talagala2018,Bauer2020} and recommendation rules~\cite{Wang2009,Zuefle2019}. In general, meta-learner after the induction phase is capable of indicating which learning method is the most appropriate to a given problem~\cite{Rokach2006}. The meta-learning concept for time-series forecasting uses a machine learning model (i.e., decision tree or ensemble of trees) and trains it on a set of features, -- various characteristics of time-series -- to recommend the most suitable univariate time-series model. It was also found that meta-learning is effective even when the meta-learner is trained on time-series from one domain and tested on time-series from another~\cite{Ali2018}, suggesting machine learning universal capability more widely known as transfer learning. FFORMS (Feature-based FORecast Model Selection)~\cite{Talagala2018} idea was implemented in R package \emph{seer} besides participation in M4 competition. However, due to mediocre performance, it was further developed into adaptive weight-producing forecasting ensemble FFORMA (Feature-based FORecast Model Averaging)~\cite{Manso2018,Manso2020}, available in R package \emph{M4metalearning}, achieving second place in M4 competition. Three novel approaches for forecasting method recommendation, where the meta-learning task was based on classification or regression or both,  were evaluated in~\cite{Bauer2020} with recommendation considering explicitly a machine learning-based regressor method instead of a statistical one. Meta-learning approach of weigh-producing nature featuring double-channel convolutional neural network was introduced in~\cite{MaFiles2021}. It outperformed FFORMA and other variants of meta-learning strategies in retail sales forecasting. However, only 2 out of 9 strategies explored were univariate. In contrast, others, including the M5 winner, required influential factors (such as price, promotions, seasonality, and calendar events) with historical values and values along the forecasting horizon.

The main difference of FFORMA and M5 approaches over FFORMS and other forecasting method recommendation systems referenced above is that weights for a pool of models in an ensemble are recommended instead of a single best model. A detailed overview of historical meta-learning approaches can be found in~\cite{MaFiles2021}. Building upon the success of FFORMA, we propose to simplify meta-learning by decomposing it into two separate regression tasks, where A1 assistant ranks the pool of potential time-series models and A2 assistant recommends ensemble size to cap the ranked list. Such simplification avoids the tedious process of weight tuning and could diminish overfitting risk due to overtuning.

\section{Methodology}
\label{sec:methodology}

\begin{figure}[!htb]
\centering
\includegraphics[width=0.99\textwidth]{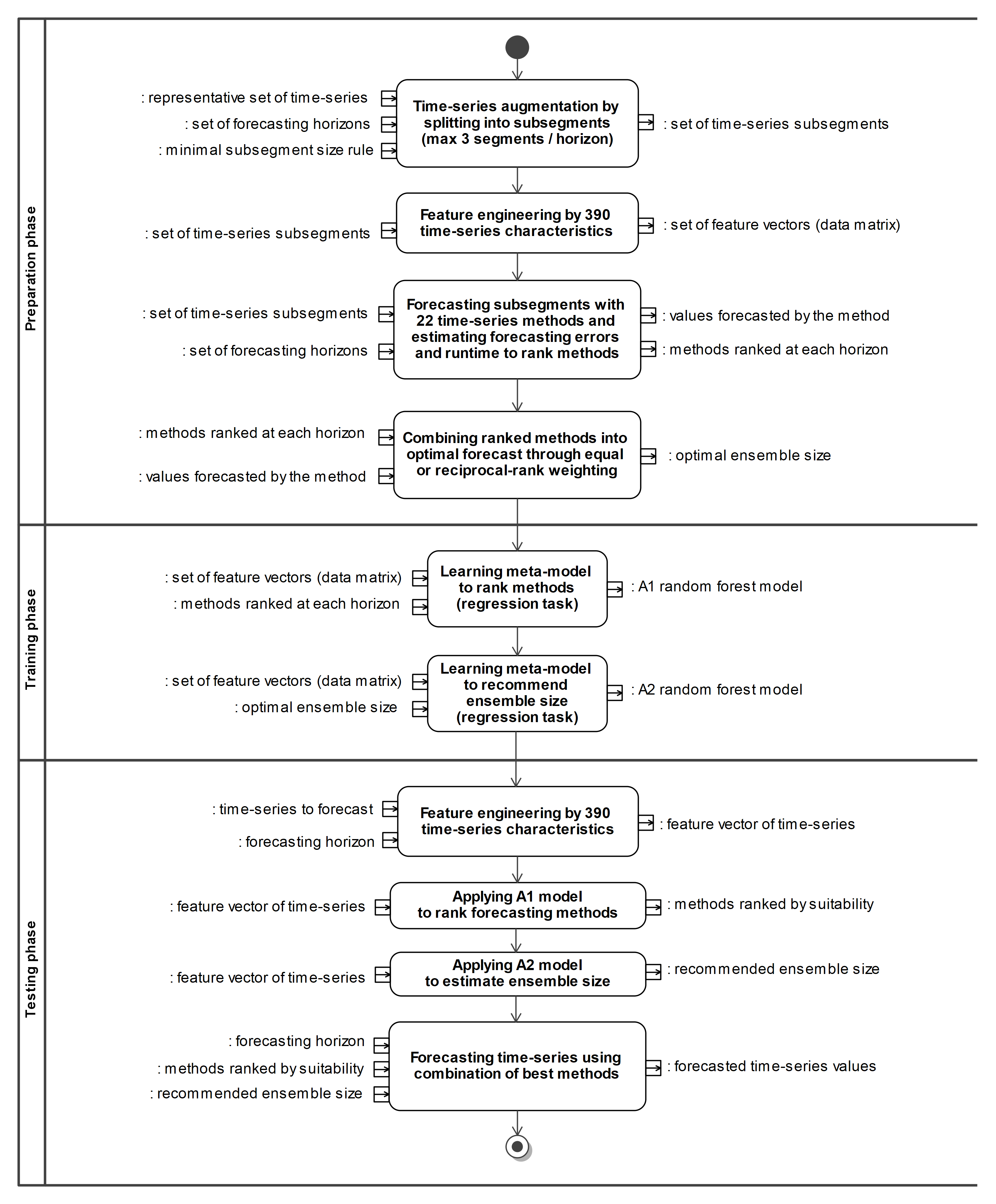}
\caption{Methodology outline by unified modeling language activity diagram: input (\emph{left hand side}) and output (\emph{right hand side}) artefacts are listed besides each action block and all process is split into 3 phases. To adhere to machine learning standards, we assure that due to performed cross-validation a set of time-series in the preparation and training phases do not mach time-series in the testing phase.}
\label{fig:UML}
\end{figure}

Assistants A1 and A2 use time-series features for modeling and meta-learning target attribute, which corresponds to a rank of a specific forecasting technique for A1 and a recommended ensemble size for A2. Each time-series used in meta-learning preparation phase is split into train/test parts. Features are calculated on the training part, while forecasts using a pool of univariate forecasting models are obtained on the testing part. Forecasting errors on the testing part are estimated, and forecasting models are ranked for A1; meanwhile, all possible ensembles are evaluated for A2. After the training phase of A1 and A2 models, their usefulness in helping with time-series forecasting is evaluated using M4-micro dataset in the testing phase. Methodology pipeline is illustrated by Fig.~\ref{fig:UML}. Additionally, variable importance from A1 and A2 models is reported in performed experiments.

\subsection{Time-series features}

\begin{table}[!htb]
\scalefont{0.9}
\centering
\caption{Overview of time-series features for meta-learning: 130 time-series characteristics in total.}
\label{tab:features}
\begin{tabular}{p{0.12\linewidth}cp{0.76\linewidth}}
\toprule
{\bf Feature set} & {\bf Size} & {\bf R package(-s) used // detailed list of   feature names} \\
\toprule
catch22 & 22 & \emph{catch22} // DN\_HistogramMode\_5, DN\_HistogramMode\_10, CO\_f1ecac,   CO\_FirstMin\_ac, CO\_HistogramAMI\_even\_2\_5, CO\_trev\_1\_num,   MD\_hrv\_classic\_pnn40, SB\_BinaryStats\_mean\_longstretch1,   SB\_TransitionMatrix\_3ac\_sumdiagcov, PD\_PeriodicityWang\_th0\_01, CO\_Embed2\_Dist\_tau\_d\_expfit\_meandiff, IN\_AutoMutualInfoStats\_40\_gaussian\_fmmi, FC\_LocalSimple\_mean1\_tauresrat,   DN\_OutlierInclude\_p\_001\_mdrmd, DN\_OutlierInclude\_n\_001\_mdrmd,   SP\_Summaries\_welch\_rect\_area\_5\_1, SB\_BinaryStats\_diff\_longstretch0,   SB\_MotifThree\_quantile\_hh, SC\_FluctAnal\_2\_rsrangefit\_50\_1\_logi\_prop\_r1,   SC\_FluctAnal\_2\_dfa\_50\_1\_2\_logi\_prop\_r1, SP\_Summaries\_welch\_rect\_centroid,   FC\_LocalSimple\_mean3\_stderr \\
\midrule
tsfeats & 13 & \emph{tsfeatures} // stability, lumpiness, crossing.points.fraction,   flat.spots.fraction, nonlinearity, ur.kpss, ur.pp, arch.lm, ACF1, ACF10.SS,   ACF.seas, PACF10.SS, PACF.seas \\
\midrule
stlfeats & 12 & \emph{tsfeatures} // nperiods, seasonal\_period, trend, spike, linearity,   curvature, e\_acf1, e\_acf10, seasonal\_strength, peak, trough, lambda \\
\midrule
hctsa & 13 & \emph{tsfeatures} // embed2\_incircle\_1, embed2\_incircle\_2, ac\_9, firstmin\_ac,   trev\_num, motiftwo\_entro3, walker\_propcross, std1st\_der,   boot\_stationarity\_fixed, boot\_stationarity\_ac2, histogram\_mode\_10,   outlierinclude\_mdrmd, first\_acf\_zero\_crossing \\
\midrule
heterogeneity, portmanteau & 6 & \emph{tsfeatures}, \emph{WeightedPortTest} // arch\_acf, garch\_acf, arch\_r2, garch\_r2,   lag1.Ljung.Box, lagF.Ljung.Box \\ \hline
stationarity, normality & 10 & \emph{tseries}, \emph{stats}, \emph{nortest} // ADF, KPSS.Level,   KPSS.Trend, PP, ShapiroWilk, Lilliefors,   AndersonDarling, Pearson, CramerVonMises, ShapiroFrancia \\
\midrule
kurtosis, skewness, misc & 11 & \emph{PerformanceAnalytics} // kurtosis.fisher, kurtosis.sample,   skewness.fisher, skewness.sample, skewness.variability, skewness.volatility,   skewness.kurtosis.ratio, misc.smoothing.index, misc.Kelly.ratio,   misc.drowpdown.average.depth, misc.drowpdown.average.length \\
\midrule
Hurst & 17 & \emph{PerformanceAnalytics}, \emph{longmemo}, \emph{tsfeatures}, \emph{liftLRD}, \emph{pracma}, \emph{fractal} // PerformanceAnalytics, Whittle, HaslettRaftery, lifting, pracma.Hs,   pracma.Hrs, pracma.He, pracma.Hal, pracma.Ht, fractal.spectral.lag.window,   fractal.spectral.wosa, fractal.spectral.multitaper, fractal.block.aggabs,   fractal.block.higuchi, fractal.ACVF.beta, fractal.ACVF.alpha, fractal.ACVF.HG \\
\midrule
fractality & 7 & \emph{fractaldim} // HallWood, DCT, wavelet, variogram, madogram, rodogram,   periodogram \\
\midrule
entropy & 9 & \emph{TSEntropies}, \emph{ForeCA} // TSE.approximate, TSE.fast.sample, TSE.fast.approx,   spectral.smoothF.wosa, spectral.smoothF.direct, spectral.smoothF.multitaper,   spectral.smoothT.wosa, spectral.smoothT.direct, spectral.smoothT.multitaper \\
\midrule
anomaly & 10 & \emph{pracma}, \emph{anomalize} // fraction.TukeyMAD, twitter.iqr.fraction,   twitter.iqr.infraction.pos, twitter.iqr.fraction.pos,   twitter.iqr.abs.last.pos, twitter.iqr.rel.last.pos,   twitter.iqr.infraction.neg, twitter.iqr.fraction.neg,   twitter.iqr.abs.last.neg, twitter.iqr.rel.last.neg \\
\bottomrule
\end{tabular}
\end{table}

Feature engineering for assistant models consisted of pooling various known time-series characteristics into a collection of 130 features (see Table~\ref{tab:features}). Almost half of all features were previously introduced as state-of-the-art time-series features, available in R packages \emph{catch22}~\cite{Lubba2019}, consisting of carefully selected 22 features, and \emph{tsfeatures}~\cite{Hyndman2019}, consisting of 42 features also used in FFORMA framework~\cite{Manso2018,Manso2020}.

Seeking to extend characteristics of time-series we considered calculating features not only on the original data (\emph{orig}), but also on the results of 2 transformations (\emph{diff} and \emph{log}):
\begin{itemize}
\item \emph{diff} - first differences help to improve/achieve stationarity;
\item \emph{log} - logarithmic transform has variance stabilizing properties.
\end{itemize}

Note that \emph{log} transformation here is not applied column-wise to the table with extracted features, but to the original time-series, which seeks to achieve a variance-stabilizing effect on the dynamics at hand. Calculating 130 features on 3 variants of time-series (\emph{orig}, \emph{diff} and \emph{log}) results in a final set of 390 features, which are later used for building assistant meta-learners.

\subsection{Forecasting models}

\begin{table}[!htb]
\caption{A selected pool of 22 base models for univariate time-series forecasting. Most models are statistical, except for NNAR and xgb, which are based on machine learning. The horizontal line separates a few simple models from the remaining complex ones.}
\label{tab:univariate}
\scalefont{0.9}
\centering
\begin{tabular}{ l l p{0.6\linewidth} }
\toprule
{\bf Model} & {\bf R package::function} & {\bf Description} \\
\toprule
SNaive & forecast::snaive & Seasonal naïve method \\
LinTrend & forecast::tslm & Linear trend \\
LinTrendSeason & forecast::tslm & Linear trend with seasonal dummies \\
QuadTrend & forecast::tslm & Quadratic trend \\
QuadTrendSeason & forecast::tslm & Quadratic trend with seasonal dummies \\
\midrule
TSB & tsintermittent::tsb & Teunter-Syntetos-Babai method (based upon Croston for intermittent demand) with optimized parameters~\cite{Kourentzes2014} \\
ARIMA & forecast::auto.arima & Autoregressive integrated moving average~\cite{HyndmanKhandakar2008} \\
SARIMA & forecast::auto.arima & Seasonal autoregressive integrated moving average~\cite{HyndmanKhandakar2008} \\
ETS & forecast::ets & Family of exponential smoothing state space models~\cite{Hyndman2002,Hyndman2008} \\
HoltWinters & stats::HoltWinters & Holt-Winters filtering with additive seasonality~\cite{Winters1960} \\
Theta & forecast::thetaf & Theta method - simple exponential smoothing with drif~\cite{Assimakopoulos2000} \\
STL-ARIMA & forecast::stlm & ARIMA model on seasonal decomposition of time series~\cite{Cleveland1990} \\
STL-ETS & forecast::stlm & ETS model on seasonal decomposition of time series~\cite{Cleveland1990} \\
StructTS & stats::StructTS & Basic stuctural model - local trend with seasonality~\cite{Durbin2012} \\
BATS & forecast::tbats & Exponential smoothing with Box-Cox transform, ARMA errors, trend and complex seasonality~\cite{Livera2011} \\
Prophet & prophet::prophet & Decomposable time series and generalized additive model with non-linear trends~\cite{Taylor2017} \\
NNAR & forecast::nnetar & Neural network with a hidden layer and lagged inputs~\cite{Hyndman2018} \\
xgb-none & forecastxgb::xgbar & Extreme gradient boosting model with lagged inputs~\cite{Ellis2016} \\
xgb-decompose & forecastxgb::xgbar & Extreme gradient boosting model with lagged inputs and decomposition-based seasonal adjustment~\cite{Ellis2016} \\
thief-ARIMA & thief::thief & Temporal hierarchical approach with ARIMA at each level~\cite{Athanasopoulos2017} \\
thief-ETS & thief::thief & Temporal hierarchical approach with ETS at each level~\cite{Athanasopoulos2017} \\
thief-Theta & thief::thief & Temporal hierarchical approach with Theta at each level~\cite{Athanasopoulos2017} \\
\bottomrule
\end{tabular}
\end{table}

A representative pool of 22 univariate time-series forecasting models was selected (see Table~\ref{tab:univariate}). The diversity of models to consider as a potential ensemble member varies from simple, such as the seasonal naive and linear trend, to complex BATS and Prophet models. However, most of them are statistical, except for machine learning approaches NNAR and xgb. Model implementations from 6 R packages were used, where parameters when creating a model on time-series training part were chosen automatically if model implementation had that capability.

\subsection{Forecasting errors}

After fitting the univariate time-series model on the training part, forecasting can be performed for a required number of steps ahead, i.e., forecasting horizon. Comparing forecasted values with ground truth allows evaluating how accurate the forecast was, and forecasting errors are used for this purpose. We estimate three absolute and three relative forecasting errors.

Absolute forecasting errors considered:
\begin{itemize}
 \item RMSE - root mean squared error;
 \item MAE - mean absolute error;
 \item MDAE - median absolute error.
\end{itemize}

Relative forecasting errors considered:
\begin{itemize}
 \item SMAPE - symmetric mean absolute percentage error;
 \item MAAPE - mean arctangent absolute percentage error~\cite{Kim2016};
 \item MASE - mean absolute scaled error~\cite{Hyndman2006}.
\end{itemize}

The final ranking of forecasting models was constructed by averaging individual rankings, obtained for each error type separately. After averaging out individual rankings, a faster model was given priority in the final ranking in case of ties. Incorporating absolute and relative errors into the final ranking allows us to sort out models more comprehensively without less bias towards a single type of error. Relative forecasting errors were also reported in experiments to compare the introduced approach to benchmark methods (Theta and Comb).

\subsection{Meta-learner model}

Meta-learner for our experiments was random forest (RF)~\cite{Breiman2001} regression machine learning model. RF is an ensemble of many (\emph{ntrees} in total) CART (classification and regression tree) instances. Each CART is built on an independent bootstrap sample of the original dataset while selecting from a random subset (of size \emph{mtry}) of features at each tree node. Fast RF implementation in R package \emph{ranger}~\cite{Wright2017} was chosen, which, conveniently for the specifics of our assistants, allows us to always include some variables as candidates for a binary node split besides \emph{mtry} randomly selected ones. Time-series features were left for random selection, but a few critical meta-information features were set to \emph{always.split.variables} parameter. Meta-information features were forecasting horizon length and data type (daily, weekly or monthly). Additionally, A1 assistant included model name (first column in Table~\ref{tab:univariate}) and three dummy indicators on model capabilities such as seasonality, complexity, and decomposition.

RF size \emph{ntrees} was fixed at 256, as recommended in literature~\cite{Oshiro2012,Probst2017}. Classical RF should be composed of unpruned CART, allowing growing trees to maximal possible depth, corresponding to \emph{min.node.size}=1 setting, but in our case \emph{min.node.size} parameter was tuned together with \emph{mtry} using Bayesian optimization in R package \emph{tuneRanger} with 21 warm-up and 9 tuning iterations. The minimization objective for A1 assistant was the out-of-bag root mean squared logarithmic error - a variant of RMSE penalizing errors at lower values and achieving that prediction of best-ranked cases is more precise than of worse-ranked candidates. The minimization objective for the A2 assistant was a simple out-of-bag RMSE metric. Both being RF regression models, A1 could be nick-named as a "ranker" whereas A2 as a "capper" due to different tasks they are dedicated to. A1 ranks 22 forecasting models to find the best candidates for the time-series at hand, while A2 tries to propose an optimal size of forecasting ensemble, i.e., a number of best-ranked models to choose for forecast pooling.

Two variants of forecast pooling, namely, Simple (arithmetic average) and Weighted (weighted average), were evaluated for meta-learning. A1 assistant was identical in both variants, but A2 was constructed separately after evaluation of cumulative pooling of forecasts from the best A1-ranked univariate time-series models, where pooling was done either with equal weights or weights derived from reciprocal rank~\cite{Aiolfi2006}.

\section{M4-micro dataset}

We excluded two monthly cases (ID=19700 and ID=19505) from an original M4 subset of 12563 micro-economic time-series due to the lack of dynamics. Among 12561 selected cases level of aggregation was as follows: 1476 daily, 112 weekly, and 10973 monthly. All selected cases were pre-processed by segmenting them properly into train/test splits at several forecasting horizons. Forecasting horizons with a varying number of steps ahead were considered: 15, 30, 90, 180, 365, and 730 days for daily data; 4, 13, 26, 52, and 104 weeks for weekly data; 6, 12, 24, 60, and 120 months for monthly data.

\begin{figure}[!htb]
\centering
\begin{subfigure}[b]{0.325\linewidth}
\centering
\includegraphics[height=0.94\textwidth]{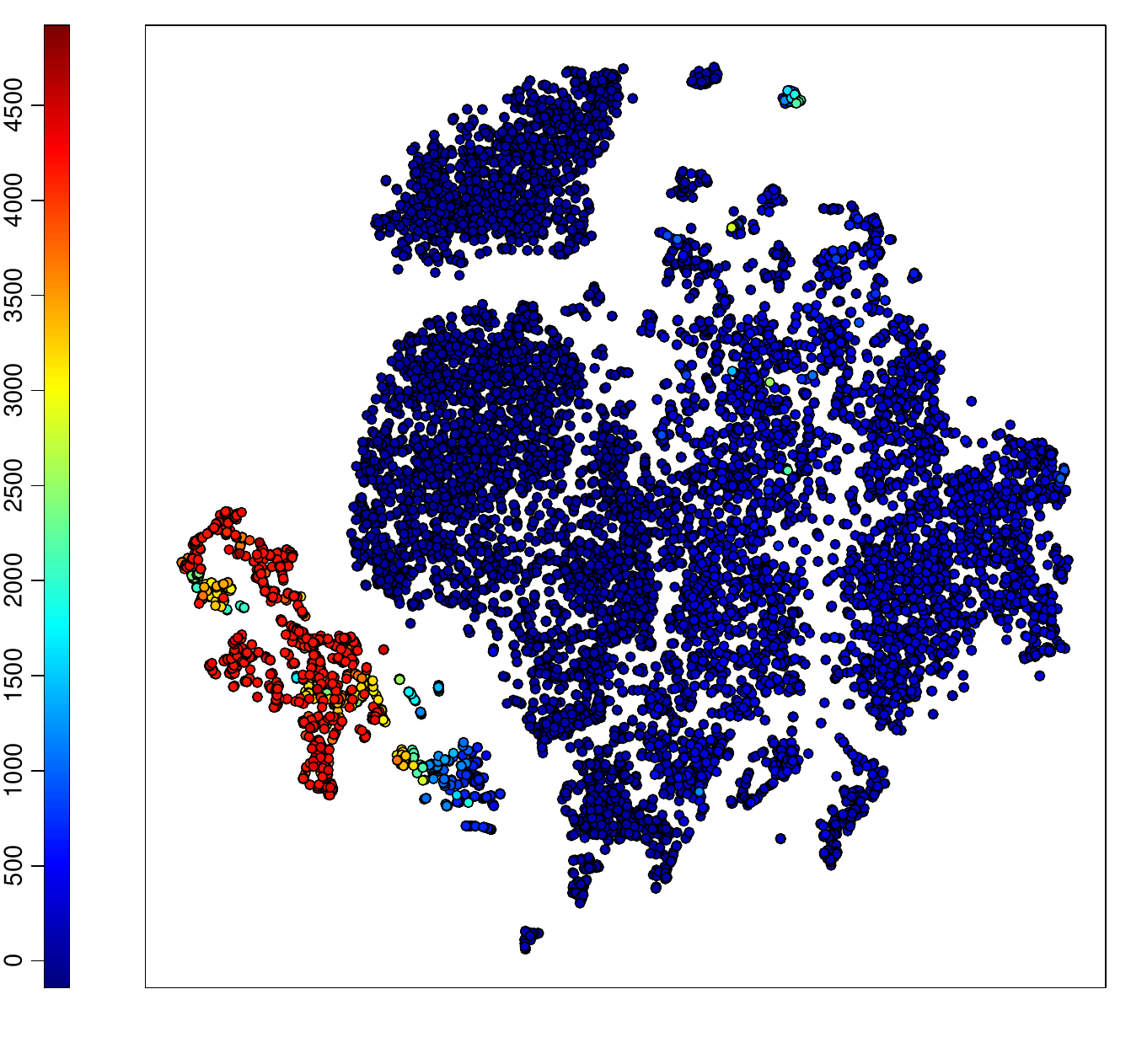}
\end{subfigure}
\thinspace
\begin{subfigure}[b]{0.325\linewidth}
\centering
\includegraphics[height=0.94\textwidth]{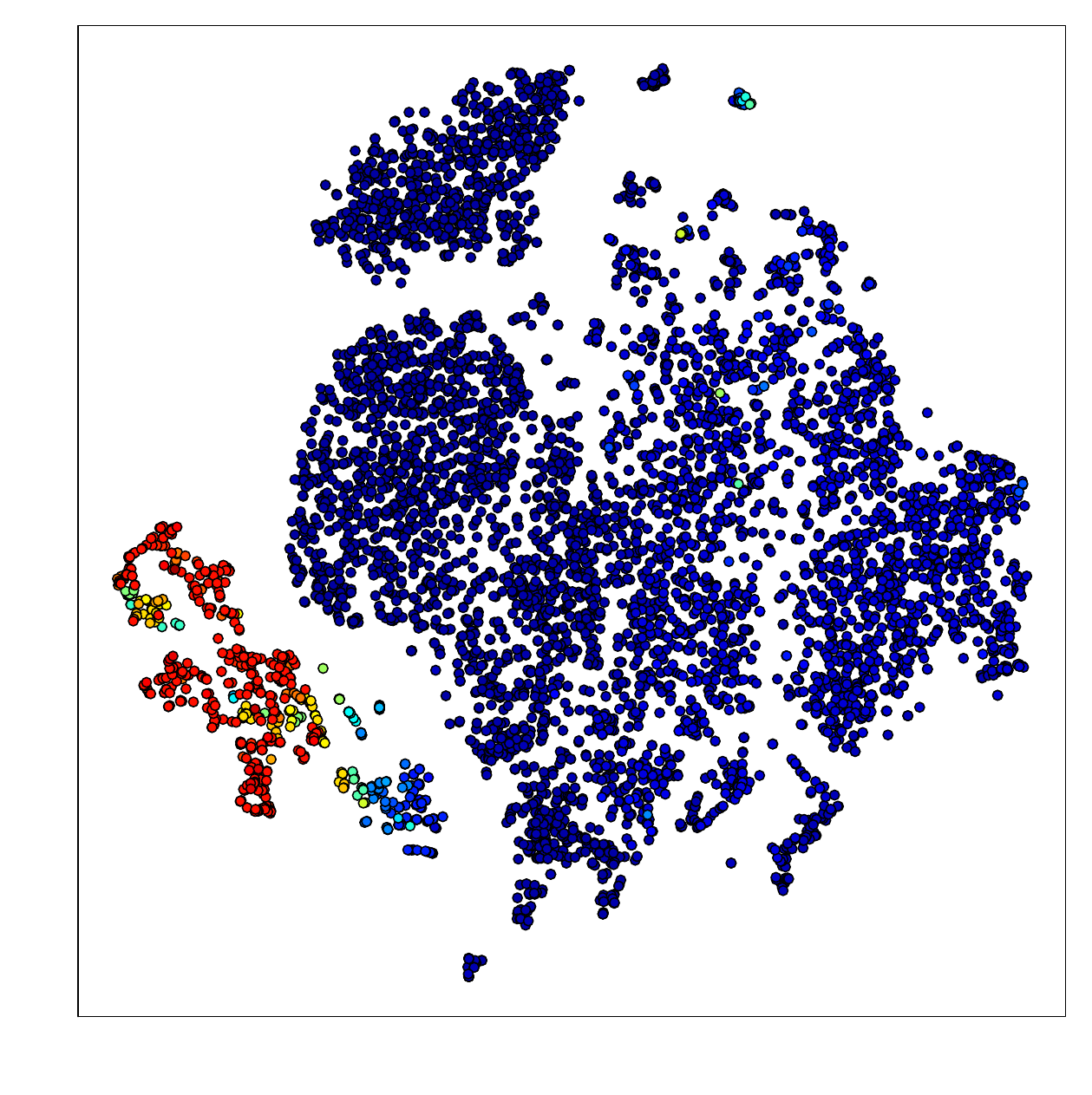}
\end{subfigure}
\begin{subfigure}[b]{0.325\linewidth}
\centering
\includegraphics[height=0.94\textwidth]{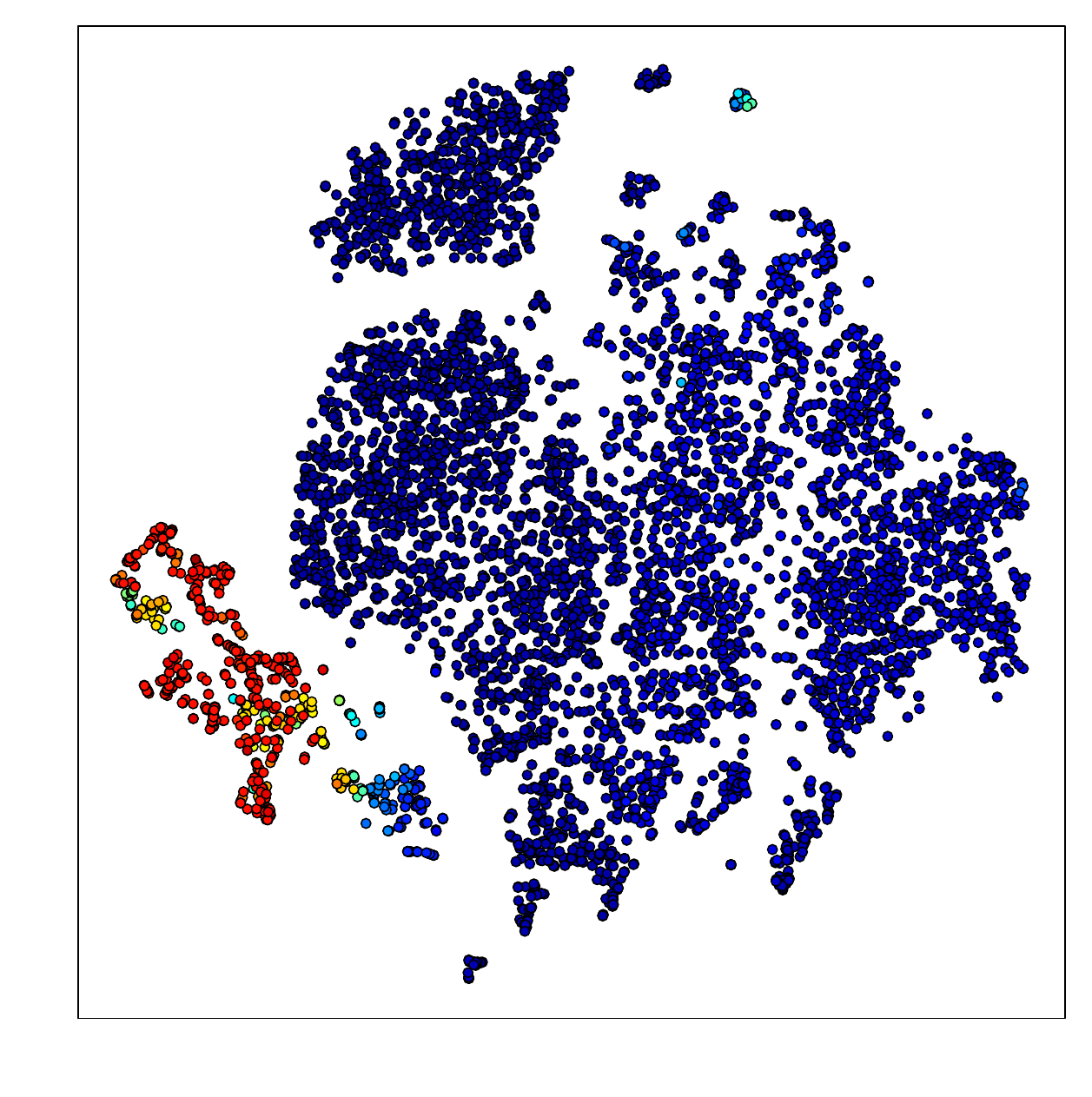}
\end{subfigure}
\caption{Visualization of M4-micro dataset by 2D $t$-SNE projection of time-series features: full sample of 12561 time-series (\emph{left}) and result after balanced sampling into 2 cross-validation folds, containing 6281 (\emph{center}) and 6280 (\emph{right}) time-series. Color of the point denotes the length of time-series. Note that to avoid any informational leakage splitting into 2 cross-validation folds was done for the original dataset before proceeding with expansions.}
\label{fig:samples}
\end{figure}

Concerned with specifics of time-series representation space~\cite{Spiliotis2020} and to avoid potential concept drift situation if meta-learners are built using random sub-spaces, we split the dataset by performing a stratified 2-fold cross-validation (2-fold CV) carefully. Stratification is done here by an efficient balanced sampling~\cite{Deville2004} on 3D $t$-SNE~\cite{Maaten2014} projection of time-series features, which allows splitting the time-series dataset into two equally-sized parts where each part covers the overall representation space of the initial dataset. The result of such stratification is visualized in Fig.~\ref{fig:samples} with resulting CV folds depicted after 2D $t$-SNE projection.

\begin{table}[!htb]
\scalefont{0.9}
\centering
\caption{Expanding M4 micro dataset for benchmarking and augmentation. The expansion was performed by segmenting each time-series into various train/test splits, fulfilling the 80/20 heuristic. Heuristic ensures that the amount of data available for time-series model training is at least four times larger than for testing. The amount of data for estimating forecasting error of the meta-learner solution is defined by the forecasting horizon. Full sample for benchmarking at the smallest forecasting horizon consists of 12561 time-series, visualized on the left of Fig.~\ref{fig:samples}.}
\label{tab:expansion}
\begin{tabular}{r|ccc|ccc}
\toprule
\multirow{2}{*}{\bf Horizon} & \multicolumn{3}{c|}{\bf Initial expansion for benchmarking} & \multicolumn{3}{c}{\bf Final expansion for augmentation} \\
 & {\bf Full sample} & {\bf CV fold 1} & {\bf CV fold 2} & {\bf Full sample} & {\bf CV fold 1} & {\bf CV fold 2} \\
 \toprule
15 days & 1476        & 736       & 740       & 4374        & 2178      & 2196      \\
30 days & 1443        & 720       & 723       & 4201        & 2100      & 2101      \\
90 days & 1335        & 667       & 668       & 3699        & 1853      & 1846      \\
180 days & 1181        & 593       & 588       & 3313        & 1661      & 1652      \\
365 days & 1047        & 524       & 523       & 2693        & 1346      & 1347      \\
730 days & 780         & 389       & 391       & 780         & 389       & 391       \\
    $\sum$ & 7262        & 3629      & 3633      & 19060       & 9527      & 9533      \\
\midrule
4 weeks & 112         & 58        & 54        & 206         & 106       & 100       \\
13 weeks & 112         & 58        & 54        & 206         & 106       & 100       \\
26 weeks & 47          & 24        & 23        & 141         & 72        & 69        \\
52 weeks & 47          & 24        & 23        & 119         & 60        & 59        \\
104 weeks & 36          & 18        & 18        & 76          & 40        & 36        \\
    $\sum$  & 354         & 182       & 172       & 748         & 384       & 364       \\
\midrule
6 months & 10973       & 5486      & 5487      & 32904       & 16452     & 16452     \\
12 months & 10973       & 5486      & 5487      & 22317       & 11166     & 11151     \\
24 months & 5574        & 2792      & 2782      & 14110       & 7055      & 7055      \\
60 months & 3432        & 1713      & 1719      & 3630        & 1813      & 1817      \\
120 months & 65          & 31        & 34        & 77          & 33        & 44        \\
    $\sum$ & 31017       & 15508     & 15509     & 73038       & 36519     & 36519     \\
\bottomrule
    $\sum$ & 38633       & 19319     & 19314     & 92846       & 46430     & 46416     \\
\bottomrule
\end{tabular}
\end{table}

Time-series expansion by segmenting data into several train/test hold-out splits was done as follows. Initially, we expand a set of time-series from 12561 to 38633 (see initial expansion in Table~\ref{tab:expansion}) to be able to test various forecasting horizons. Then we increase the amount of time-series from 38633 to 92846 (see final expansion in Table~\ref{tab:expansion}) by considering additional splitting time-series in half to be able to train meta-learners on more data. Initial expansion was carried out to benchmark various forecasting horizons, and only recent data was split-off for testing. In contrast, final expansion was considered as a way of data augmentation to harvest more training data for meta-learners. Besides initial expansion, extra two splits were done where possible - hold-out on older and newer halves of time-series. Following the recommendation of~\cite{Cerqueira2020} for using an out-of-sample hold-out split in multiple testing periods, we consider 80/20 as a sufficient train/test ratio. After leaving out the last observations for testing (based on the forecasting horizon's length), if the amount of the training data drops down below 80\% we refuse to segment time-series.

\section{Experimental results}

The forecasting experiment was performed using 2-fold CV by creating assistants on CV fold 1 of final expansion and testing success of assistant-recommended forecasting ensemble on CV fold 2 of initial expansion and vice versa. Besides forecasting using the proposed approach, benchmark methods Theta and Comb were used on initial expansion, and relative forecasting errors were calculated for comparison.

\begin{table}[!htb]
\caption{Forecasting results according to SMAPE forecasting error. $\sum$ denotes results over all horizons, the best result for each row is formatted in italic-bold.}
\label{tab:SMAPE}
\scalefont{0.9}
\centering
\begin{tabular}{r|rrrr}
\toprule
{\bf Horizon} & \multicolumn{1}{c}{\bf Theta} & \multicolumn{1}{c}{\bf Comb} & \multicolumn{1}{c}{\bf Simple} & \multicolumn{1}{c}{\bf Weighted} \\
\toprule
15 days &  2.507 $\pm$  0.152 &  2.511 $\pm$  0.146 & {\bf \emph{ 2.326 $\pm$  0.145}} &  2.332 $\pm$  0.153 \\ 
30 days &  3.354 $\pm$  0.277 &  3.340 $\pm$  0.282 &  3.233 $\pm$  0.287 & {\bf \emph{ 3.229 $\pm$  0.286}} \\ 
90 days &  5.587 $\pm$  0.312 &  5.576 $\pm$  0.318 & {\bf \emph{ 5.060 $\pm$  0.310}} &  5.064 $\pm$  0.310 \\ 
180 days &  8.296 $\pm$  0.444 &  8.565 $\pm$  0.452 &  7.298 $\pm$  0.438 & {\bf \emph{ 7.282 $\pm$  0.438}} \\ 
365 days & 17.258 $\pm$  0.919 & 17.304 $\pm$  0.895 & {\bf \emph{14.098 $\pm$  0.829}} & 14.145 $\pm$  0.828 \\ 
730 days & 18.270 $\pm$  1.003 & 19.052 $\pm$  1.042 & 16.396 $\pm$  0.974 & {\bf \emph{16.395 $\pm$  0.991}} \\ 
$\sum$ &  8.003 $\pm$  0.245 &  8.133 $\pm$  0.248 & {\bf \emph{ 7.026 $\pm$  0.225}} &  7.031 $\pm$  0.226 \\
\midrule
4 weeks &  9.483 $\pm$  1.152 &  9.653 $\pm$  1.198 &  7.941 $\pm$  1.024 & {\bf \emph{ 7.867 $\pm$  1.000}} \\ 
13 weeks &  9.365 $\pm$  1.130 &  8.910 $\pm$  1.047 & {\bf \emph{ 8.481 $\pm$  1.076}} &  8.637 $\pm$  1.222 \\ 
26 weeks &  8.895 $\pm$  4.395 &  8.594 $\pm$  4.404 &  8.184 $\pm$  4.378 & {\bf \emph{ 8.092 $\pm$  4.371}} \\ 
52 weeks & 13.340 $\pm$  4.619 & 12.575 $\pm$  4.036 & {\bf \emph{10.763 $\pm$  3.519}} & 10.922 $\pm$  3.784 \\ 
104 weeks & 17.452 $\pm$  5.277 & 18.198 $\pm$  6.037 & 16.273 $\pm$  4.767 & {\bf \emph{16.008 $\pm$  4.855}} \\ 
$\sum$ & 10.690 $\pm$  1.126 & 10.534 $\pm$  1.127 & {\bf \emph{ 9.366 $\pm$  1.010}} &  9.374 $\pm$  1.041 \\
\midrule 
6 months & 12.150 $\pm$  0.278 & 12.147 $\pm$  0.286 &  9.158 $\pm$  0.214 & {\bf \emph{ 9.096 $\pm$  0.212}} \\ 
12 months & 12.183 $\pm$  0.246 & 12.874 $\pm$  0.269 & 10.479 $\pm$  0.222 & {\bf \emph{10.420 $\pm$  0.220}} \\ 
24 months &  9.544 $\pm$  0.352 &  9.325 $\pm$  0.353 & {\bf \emph{ 8.451 $\pm$  0.327}} &  8.486 $\pm$  0.333 \\ 
60 months & 12.724 $\pm$  0.604 & 14.780 $\pm$  0.740 & 11.617 $\pm$  0.583 & {\bf \emph{11.449 $\pm$  0.567}} \\ 
120 months & 15.051 $\pm$  5.820 & 13.656 $\pm$  4.244 & 10.798 $\pm$  3.371 & {\bf \emph{10.650 $\pm$  3.329}} \\ 
$\sum$ & 11.763 $\pm$  0.161 & 12.192 $\pm$  0.174 &  9.774 $\pm$  0.140 & {\bf \emph{ 9.719 $\pm$  0.139}} \\
\bottomrule
$\sum$ & 11.047 $\pm$  0.139 & 11.414 $\pm$  0.149 &  9.254 $\pm$  0.121 & {\bf \emph{ 9.210 $\pm$  0.120}} \\
\bottomrule
\end{tabular}
\end{table}

Results by SMAPE (see Table~\ref{tab:SMAPE}) demonstrate that both Simple and Weighted variants of meta-learning outperform Theta and Comb techniques. Weighted slightly outperformed Simple variant for more than half (10 out of 16) horizons and also overall (see the last row in Table~\ref{tab:SMAPE}).

\begin{table}[!htb]
\caption{Forecasting results according to MAAPE forecasting error. $\sum$ denotes results over all horizons, the best result for each row is formatted in italic-bold.}
\label{tab:MAAPE}
\scalefont{0.9}
\centering
\begin{tabular}{r|rrrr}
\toprule
{\bf Horizon} & \multicolumn{1}{c}{\bf Theta} & \multicolumn{1}{c}{\bf Comb} & \multicolumn{1}{c}{\bf Simple} & \multicolumn{1}{c}{\bf Weighted} \\
\toprule
15 days &  2.520 $\pm$  0.151 &  2.536 $\pm$  0.153 &  2.335 $\pm$  0.148 & {\bf \emph{ 2.335 $\pm$  0.153}} \\ 
30 days &  3.312 $\pm$  0.268 &  3.296 $\pm$  0.269 &  3.185 $\pm$  0.273 & {\bf \emph{ 3.183 $\pm$  0.272}} \\ 
90 days &  5.538 $\pm$  0.316 &  5.526 $\pm$  0.318 & {\bf \emph{ 5.033 $\pm$  0.317}} &  5.036 $\pm$  0.317 \\ 
180 days &  8.047 $\pm$  0.433 &  8.366 $\pm$  0.444 &  7.161 $\pm$  0.434 & {\bf \emph{ 7.142 $\pm$  0.434}} \\ 
365 days & 15.431 $\pm$  0.783 & 15.538 $\pm$  0.772 & {\bf \emph{13.132 $\pm$  0.766}} & 13.165 $\pm$  0.765 \\ 
730 days & 17.696 $\pm$  0.971 & 17.989 $\pm$  0.975 & 16.306 $\pm$  0.993 & {\bf \emph{16.286 $\pm$  1.000}} \\ 
$\sum$ &  7.622 $\pm$  0.226 &  7.719 $\pm$  0.228 &  6.842 $\pm$  0.218 & {\bf \emph{ 6.842 $\pm$  0.219}} \\
\midrule
4 weeks &  8.750 $\pm$  1.027 &  8.856 $\pm$  1.061 &  7.570 $\pm$  0.953 & {\bf \emph{ 7.514 $\pm$  0.938}} \\ 
13 weeks &  9.475 $\pm$  1.146 &  8.896 $\pm$  1.057 & {\bf \emph{ 8.636 $\pm$  1.142}} &  8.717 $\pm$  1.207 \\ 
26 weeks &  8.471 $\pm$  4.175 &  8.233 $\pm$  4.190 &  7.945 $\pm$  4.213 & {\bf \emph{ 7.849 $\pm$  4.208}} \\ 
52 weeks & 12.084 $\pm$  3.451 & 11.620 $\pm$  3.374 & {\bf \emph{10.231 $\pm$  3.223}} & 10.270 $\pm$  3.320 \\ 
104 weeks & 15.683 $\pm$  3.885 & 16.119 $\pm$  4.218 & 15.251 $\pm$  3.890 & {\bf \emph{14.979 $\pm$  3.921}} \\ 
$\sum$ & 10.090 $\pm$  0.955 &  9.891 $\pm$  0.962 &  9.091 $\pm$  0.938 & {\bf \emph{ 9.064 $\pm$  0.949}} \\
\midrule
6 months & 11.059 $\pm$  0.242 & 10.983 $\pm$  0.243 &  8.892 $\pm$  0.207 & {\bf \emph{ 8.840 $\pm$  0.206}} \\ 
12 months & 12.239 $\pm$  0.244 & 13.203 $\pm$  0.273 & 10.648 $\pm$  0.224 & {\bf \emph{10.593 $\pm$  0.223}} \\ 
24 months &  9.119 $\pm$  0.320 &  8.906 $\pm$  0.319 & {\bf \emph{ 8.190 $\pm$  0.306}} &  8.200 $\pm$  0.307 \\ 
60 months & 12.500 $\pm$  0.601 & 13.512 $\pm$  0.601 & 11.330 $\pm$  0.556 & {\bf \emph{11.197 $\pm$  0.548}} \\ 
120 months & 11.883 $\pm$  3.194 & 11.499 $\pm$  2.941 &  9.389 $\pm$  2.538 & {\bf \emph{ 9.341 $\pm$  2.539}} \\ 
$\sum$ & 11.289 $\pm$  0.151 & 11.676 $\pm$  0.158 &  9.658 $\pm$  0.136 & {\bf \emph{ 9.607 $\pm$  0.136}} \\
\bottomrule
$\sum$ & 10.589 $\pm$  0.129 & 10.916 $\pm$  0.135 &  9.123 $\pm$  0.118 & {\bf \emph{ 9.082 $\pm$  0.117}} \\
\bottomrule
\end{tabular}
\end{table}

Results by MAAPE (see Table~\ref{tab:MAAPE}) demonstrate that both Simple and Weighted variants of meta-learning outperform Theta and Comb techniques. Weighted slightly outperformed Simple variant for more than half (11 out of 16) horizons and also overall (see the last row in Table~\ref{tab:MAAPE}).

\begin{table}[!htb]
\scalefont{0.9}
\centering
\caption{Forecasting results according to MASE forecasting error. $\sum$ denotes results over all horizons, the best result for each row is formatted in italic-bold.}
\label{tab:MASE}
\begin{tabular}{r|rrrr}
\toprule
{\bf Horizon} & \multicolumn{1}{c}{\bf Theta} & \multicolumn{1}{c}{\bf Comb} & \multicolumn{1}{c}{\bf Simple} & \multicolumn{1}{c}{\bf Weighted} \\ \toprule
115 days &  1.047 $\pm$  0.050 &  1.056 $\pm$  0.055 & {\bf \emph{ 0.963 $\pm$  0.046}} &  0.964 $\pm$  0.049 \\ 
30 days &  1.365 $\pm$  0.076 &  1.362 $\pm$  0.077 &  1.311 $\pm$  0.081 & {\bf \emph{ 1.310 $\pm$  0.081}} \\ 
90 days &  2.245 $\pm$  0.106 &  2.239 $\pm$  0.107 &  1.997 $\pm$  0.102 & {\bf \emph{ 1.996 $\pm$  0.101}} \\ 
180 days &  3.265 $\pm$  0.200 &  3.334 $\pm$  0.200 &  2.910 $\pm$  0.202 & {\bf \emph{ 2.904 $\pm$  0.202}} \\ 
365 days &  5.987 $\pm$  0.309 &  6.132 $\pm$  0.320 & {\bf \emph{ 5.053 $\pm$  0.301}} &  5.060 $\pm$  0.300 \\ 
730 days &  7.676 $\pm$  0.521 &  7.798 $\pm$  0.517 & {\bf \emph{ 6.747 $\pm$  0.484}} &  6.761 $\pm$  0.504 \\ 
$\sum$ &  3.115 $\pm$  0.098 &  3.161 $\pm$  0.099 & {\bf \emph{ 2.750 $\pm$  0.091}} &  2.751 $\pm$  0.092 \\
\midrule
4 weeks &  0.579 $\pm$  0.077 &  0.592 $\pm$  0.084 &  0.480 $\pm$  0.070 & {\bf \emph{ 0.475 $\pm$  0.068}} \\ 
13 weeks &  0.486 $\pm$  0.052 &  0.476 $\pm$  0.058 & {\bf \emph{ 0.426 $\pm$  0.047}} &  0.432 $\pm$  0.051 \\ 
26 weeks &  0.681 $\pm$  0.324 &  0.680 $\pm$  0.350 &  0.622 $\pm$  0.347 & {\bf \emph{ 0.599 $\pm$  0.340}} \\ 
52 weeks &  0.993 $\pm$  0.299 &  0.967 $\pm$  0.308 &  0.821 $\pm$  0.282 & {\bf \emph{ 0.806 $\pm$  0.286}} \\ 
104 weeks &  1.713 $\pm$  0.731 &  1.689 $\pm$  0.701 & {\bf \emph{ 1.564 $\pm$  0.729}} &  1.587 $\pm$  0.753 \\ 
$\sum$ &  0.733 $\pm$  0.103 &  0.728 $\pm$  0.103 &  0.637 $\pm$  0.101 & {\bf \emph{ 0.635 $\pm$  0.103}} \\
\midrule
6 months &  0.642 $\pm$  0.011 &  0.637 $\pm$  0.011 &  0.513 $\pm$  0.010 & {\bf \emph{ 0.511 $\pm$  0.010}} \\ 
12 months &  0.730 $\pm$  0.013 &  0.750 $\pm$  0.014 &  0.625 $\pm$  0.012 & {\bf \emph{ 0.622 $\pm$  0.012}} \\ 
24 months &  1.153 $\pm$  0.029 &  1.118 $\pm$  0.029 &  0.977 $\pm$  0.027 & {\bf \emph{ 0.977 $\pm$  0.027}} \\ 
60 months &  1.986 $\pm$  0.099 &  2.428 $\pm$  0.131 &  1.919 $\pm$  0.221 & {\bf \emph{ 1.827 $\pm$  0.130}} \\ 
120 months &  4.217 $\pm$  1.153 &  4.025 $\pm$  1.120 &  3.162 $\pm$  0.982 & {\bf \emph{ 3.144 $\pm$  0.989}} \\ 
$\sum$ &  0.921 $\pm$  0.015 &  0.968 $\pm$  0.018 &  0.797 $\pm$  0.026 & {\bf \emph{ 0.785 $\pm$  0.017}} \\
\bottomrule
$\sum$ &  1.332 $\pm$  0.023 &  1.378 $\pm$  0.025 &  1.163 $\pm$  0.028 & {\bf \emph{ 1.153 $\pm$  0.023}} \\
\bottomrule
\end{tabular}
\end{table}

Results by MASE (see Table~\ref{tab:MASE}) demonstrate that both Simple and Weighted variants of meta-learning outperform Theta and Comb techniques. Weighted slightly outperformed Simple variant for more than half (12 out of 16) horizons and also overall (see the last row in Table~\ref{tab:MAAPE}).

Forecasting errors showed an expected and consistent tendency to increase together with increasing forecasting horizon length. Interestingly, MASE errors were the lowest overall for weekly whereas SMAPE and MAAPE for daily data. To summarize over all data types and horizons: among benchmark methods, Theta tends to outperform Comb slightly, and meta-learning approaches win over both benchmarks with the Weighted variant as the best.

\begin{figure*}[!htb]
\centering
\includegraphics[width=0.94\textwidth]{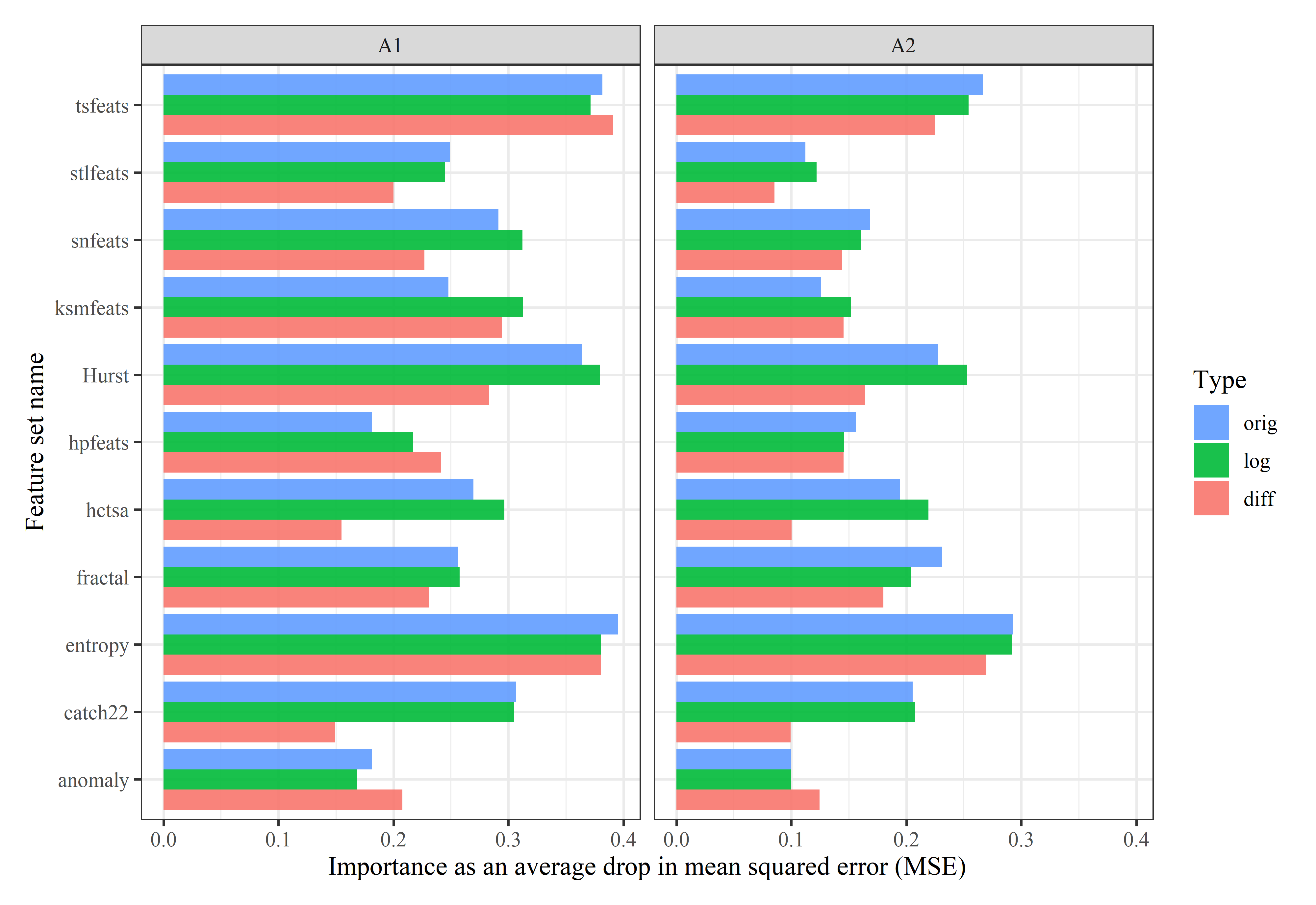}
\caption{Permutation-based variable importance from meta-learner -- a random forest regression model: A1 (\emph{left}) and A2 (\emph{right}) assistants.}
\label{fig:varImp}
\end{figure*}

We have measured the prognostic usefulness of time-series characteristics for additional insights into meta-learners by calculating permutation-based variable importance for a random forest model built on a full M4-micro dataset. Variable importance was measured as a drop in mean squared error of the random forest regression model after permuting each feature, i.e., how strongly distorting ties between a feature in question and a target affects the model's accuracy. Feature sets had several features, so their corresponding individual importance estimates were averaged for a fair comparison and are reported in Fig.~\ref{fig:varImp}. From comparison it can be noticed that both transformations (\emph{log} and \emph{diff}) besides original time-series were useful, with the exception of \emph{diff} transformation for \emph{hctsa} and \emph{catch22} features. Interestingly, the pre-processing of time-series by \emph{log} transformation was highly useful for most feature sets. The best feature sets for meta-learning tasks were \emph{entropy}, \emph{tsfeats}, and \emph{Hurst}. Among the least useful feature sets \emph{anomaly} could be considered for removal.

Since we did not use the full M4 dataset (just 12.561\% from the entire dataset of 100000 time-series) and tested much more forecasting horizons, it would be improper to compare SMAPE and MASE errors directly to M4 competition results. Nonetheless, the measure "\% improvement of method over the benchmark" from Table 4 in \cite{Makridakis2020} could lend itself for comparison with the winner of M4 competition - Smyl (Uber) reached an improvement of 9.4\% in SMAPE and 7.7\% in MASE over baseline Comb method. Our Weighted variant here demonstrated an improvement of 19.3\% in SMAPE and 16.3\% in MASE.

\section{Conclusions}

Extensive evaluation of the proposed meta-learning approach on micro-economic time-series from M4 competition demonstrated that meta-learning could outperform benchmark methods Theta and Comb. The best performance was achieved by pooling forecasts from assistant-recommended univariate time-series models using weighted average with weights corresponding to reciprocal rank. Lower forecasting errors were obtained using a weighted variant of forecasting ensemble over the Theta method: 9.21\% versus 11.05\% by SMAPE, 9.08\% versus 10.59\% by MAAPE, 1.15 versus 1.33 by MASE. The regression meta-learner model was more successful and had a better out-of-bag fit for A1 than for A2 assistant. All transformations were found to be useful for feature engineering and the most effective time-series characteristics for meta-learner were \emph{entropy}, \emph{tsfeats}, and \emph{Hurst} feature sets. Considering a more extensive set of time-series data for meta-learning and exploring automatic feature engineering's usefulness using sequence-to-sequence auto-encoder would be exciting directions for further research.

\section*{Acknowledgments}
This study contains research results from European Union funded project No. J05-LVPA-K-04-0004 "Artificial intelligence and statistical methods based time series forecasting and management" where a grant was administered by Lithuanian business promotion agency.

\section*{Data Availability Statement}
This study was a re-analysis of existing data, which is publicly available in:
\begin{itemize}
\item R packages \emph{\href{https://cran.r-project.org/package=M4comp}{M4comp}} and \emph{\href{https://robjhyndman.com/hyndsight/m4comp2018}{M4comp2018}}
\item \href{https://www.kaggle.com/yogesh94/m4-forecasting-competition-dataset}{Kaggle platform}
\item Zenodo platform:
\begin{itemize}
\item \href{http://doi.org/10.5281/zenodo.3898361}{M4 Daily Dataset}~\cite{Godahewa2020d}
\item \href{http://doi.org/10.5281/zenodo.3898376}{M4 Weekly Dataset}~\cite{Godahewa2020w}
\item \href{http://doi.org/10.5281/zenodo.3898380}{M4 Monthly Dataset}~\cite{Godahewa2020m}
\end{itemize}
\end{itemize}

\bibliographystyle{apalike}
\bibliography{references}

\end{document}